\documentclass{article}
\usepackage{spconf,amsmath,epsfig,graphicx}
\usepackage[utf8]{inputenc}
\usepackage[T1]{fontenc}
\usepackage{graphicx}
\graphicspath{{images/}}

\title{Improving performance of aircraft detection in satellite imagery while limiting the labelling effort: hybrid active learning.}

\name{Julie Imbert, Gohar Dashyan, Alex Goupilleau, Tugdual Ceillier, and Marie-Caroline Corbineau}

\address{Preligens (ex-Earthcube), Paris, France}

\begin{document}
%\ninept
%
\maketitle
\begin{abstract}

% Max 150 words
The earth observation industry provides satellite imagery with high spatial resolution and short revisit time. To allow efficient operational employment of these images, automating certain tasks has become necessary. In the defense domain, aircraft detection on satellite imagery is a valuable tool for analysts. Obtaining high performance detectors on such a task can only be achieved by leveraging deep learning and thus using a large amount of labeled data. To obtain labels of a high enough quality, the knowledge of military experts is needed. We propose a hybrid clustering active learning method to select the most relevant data to label, thus limiting the amount of data required and further improving the performances. It combines diversity- and uncertainty-based active learning selection methods. For aircraft detection by segmentation, we show that this method can provide better or competitive results compared to other active learning methods.
\end{abstract} 
\section{Introduction}
\label{introduction}

Despite impressive improvements in the last decade, especially thanks to deep learning contributions, object detection remains a challenging task. Its use in the defense industry especially calls for the most reliable and efficient solutions, that must be updated regularly. This implies that, at an operational scale, neural networks should be updated regularly to ensure high performances at all times. 
Aircraft detection in satellite imagery is a use-case of deep neural networks particularly useful for analysts. In order to automate this task while reaching high operational performances, a concurrent approach with both a segmentation and a detection convolutional networks has been proposed in \cite{segretina}. Each network has a dedicated optimized architecture and training process. For training, a comprehensive, optimized and large dataset must be provided. Indeed, performances of such a framework rely heavily on the amount and quality of the data labeled by domain experts. As satellite images are huge in terms of square kilometers and as aircraft models are numerous and very diverse, aircraft identification on satellite imagery is intrinsically a difficult and time-consuming labelling task.
 
 In the scope of this work, active learning aims at selecting a given amount of data, the labelling budget, from a pool of unlabelled data with a given strategy in order to improve an existing detector as much as possible while limiting the labelling effort. This reduces both time and cost of model updates. This is especially relevant when the labelling capacities are limited despite having a large amount of data available, which is often the case in defense organisations. The data selection strategy relies on the evaluation of the informativeness of the data, here tiles of satellite images. Informativeness can for instance be quantified by the initial detector uncertainty or with respect to the diversity of a whole pool of data.
 
An uncertainty-based method is proposed in \cite{gal_deep_2017}. It uses Bayesian dropout, more specifically referred to as MC (Monte-Carlo) dropout by the authors, at inference, to obtain various predictions with a single model on the same data. These predictions allow to quantify the uncertainty of the model and to select the most uncertain data to label for the model update. Other uncertainty methods include using an ensemble of classifiers \cite{beluch_power_2018}, or using adversarial examples \cite{ducoffe_adversarial_2018}.

With uncertainty-based methods, the selected data may be redundant. To avoid this issue, methods based on the diversity of the data can rather be used. The aim is to obtain a subset of data that is as representative as possible of the whole unlabelled data pool. In \cite{sener_active_2017}, the authors solve a geometric K-Center problem by using the L2 distance between activations of the last dense layer in a CNN. They show that this Core-Set method can outperform both random selection and the uncertainty method of \cite{gal_deep_2017}. An example of a different diversity strategy presented in \cite{gissin_discriminative_2019}, uses a generator to generate a sub-dataset which is representative of the whole set of data.

Some authors propose to combine both uncertainty and diversity in a single solution. In \cite{ash_deep_2020}, uncertainty is evaluated considering the magnitude of the loss gradients in the last layer of the network. These gradients are also used to compute the diversity: a subset of data is diverse if the gradients of the examples have complementary directions. To compute the loss gradients, model predictions are used instead of ground truths. The authors show that this hybrid approach almost always outperforms uncertainty or diversity approaches.

In \cite{yoo_learning_2019}, the authors propose to learn to predict the loss value during training assuming that it is linked to the quantity of information that a single data example brings. Active learning can also be treated as a reinforcement learning task \cite{casanova_reinforced_2020}.

In this paper, we propose a novel hybrid active learning method, combining the MC dropout \cite{gal_deep_2017} and the Core-Set \cite{sener_active_2017} approaches applied to the problem of aircraft detection in satellite imagery. The main contributions are the following:
(i) a novel hybrid active learning method combining MC dropout and Core-Set methods with two variants: a naive one and a clustering one; 
(ii) a comprehensive comparison of the results of these methods with several other labelling and training strategies including two baselines and the MC dropout and Core-Set methods alone. This comparison is performed by extending the work already presented in \cite{caid} for the aircraft detection on satellite imagery by segmentation.

We first introduce the hybrid method in Section~\ref{methods}. All relevant data, metrics and performance evaluation, as well as the results, are presented in Section~\ref{results}.

\section{Method}
\label{methods}

We explore the combination of MC dropout and Core-Set strategies in a hybrid method in order to optimize the performances of an aircraft detector while having a fixed budget for data labelling. In this work, aircraft detection is performed thanks to a binary segmentation network whose output of interest is a segmentation map that gives a probability for each pixel to be part of an aircraft.

\subsection{Tile pre-selection}

As in \cite{caid}, in order to use satellite images with deep networks, we first split these images into 512x512 pixels tiles. Considering the amount of unlabelled data that may be available to update an existing detector, processing all created tiles with active learning techniques may prove to be unpractical. To further improve efficiency of the proposed method, we perform a first preliminary tile selection. The initial network considered is used to make a prediction on all available tiles. The tiles are then ordered according to the average intensity of their segmentation map for the object of interest. Tiles with a very low response have indeed proven to be almost never selected by active learning techniques. In this work, the number of pre-selected tiles is set to 5\% of all the available tiles which corresponds to around 400k tiles as seen in Table~\ref{tab:initial_models}.

\subsection{MC dropout and Core-Set for segmentation}

In \cite{caid} we have adapted MC dropout and Core-Set methods for segmentation. 

The first one, which is derived from \cite{gal_deep_2017} consists in predicting 10 times on a given tile while activating dropout in the network, using the same dropout rate as during training. An estimation of the uncertainty on the tile considering the 10 segmentation maps is then derived by (i) computing for each pixel the variance of the 10 values, (ii) computing the average of all the variances.

The second approach is inspired from \cite{sener_active_2017}. To derive reasonably sized vectors from the features extracted by the network, the feature map at the end of the decoder of the segmentation architecture, which has a size of 128x128x128 (width, height, and number of filters) for the network presented in Section~\ref{results}, is selected. Max pooling is then used to get an 8x8x128 matrix and average pooling allows to compress it into a 1x1x128 matrix, interpreted as a 1D vector. With a labelling budget of $k$, the Robust k-center algorithm \cite{sener_active_2017} finally allows to select the $k$ most representative tiles.

\subsection{Hybrid active learning for segmentation}

Uncertainty and diversity are both valuable information and active learning strategies using respectively each of them may prove to be complementary. Using uncertainty allows to sample specific tiles that have rich information for a given detector. Nevertheless, relying only on an uncertainty criterion to select data can lead to poor diversity, as the uncertainty approach does not prevent selected tiles from being similar to each other. On the other hand, methods based on diversity are intrinsically designed to overcome this problem by constituting a sub-dataset that is as representative as possible of the initial pool. In this case, hard examples, that can be very useful to improve a given model, can however be missed. Therefore, we propose a strategy to select a subset of the unlabelled pool that contains tiles that are both uncertain and representative of the whole pool. Two methods are designed:
\begin{itemize}
    \item \textbf{A naive hybrid method} - The Core-Set and MC Dropout methods are applied sequentially. Half of the labelling budget is dedicated to the Core-Set method and the other half to the MC dropout method. Tiles selected by the Core-Set method are removed from the pool before applying the MC dropout method.
    \item\textbf{A hybrid clustering method} - First, a clustering algorithm is applied in order to regroup the tiles by similarity in the feature space, using the L2 norm. Then, the most uncertain tile in each group is chosen. The number of groups is set to be equal to the labelling budget. For the implementation, we used the scikit-learn library’s method called \emph{Agglomerative Clustering} followed by the uncertainty computation using the MC dropout method.
\end{itemize}

\section{Experimental results}
\label{results}

 We extend the work we presented in \cite{caid} with two variants of a novel hybrid active learning method. For this reason, we use the same experimental protocol.

\subsection{Initial segmentation models}

In order to thoroughly compare the novel hybrid method variants to the previously tested active learning methods in \cite{caid}, we use two  segmentation models that share the same modified U-Net architecture \cite{ronneberger_u-net_2015}, but that are trained on different data:
\begin{itemize}
    \item the \emph{weak} model is trained on relatively few images. It corresponds to a model still in development that should get better performance as fast as possible;
    \item the \emph{strong} model is trained on much more data and reaches correct performances. It is representative of a model deployed in production that should be improved with time.
\end{itemize}

To design the two training datasets of different sizes, full-sized satellite images are selected based on their date of acquisition. This allows to mimic a real-life operational situation where images are gradually available in time. The images used are from \emph{Maxar} satellites - WorldView-1, -2 and -3 and GeoEye-1. For the \emph{weak} model, images acquired between 2010 and 2012 are selected. For the \emph{strong} model, the second limit is extended until 2017. These images are divided into 512x512 tiles with a resolution of 30 cm. The tiles containing aircraft are first selected and 10\% of the number of these positive tiles is randomly sampled among negative tiles to be added to the training dataset. Table~\ref{tab:initial_models} summarises the data statistics of the datasets. Each model is trained on its dedicated dataset with Adam optimizer and a weighted cross-entropy loss. The initial learning rate (LR) is 0.001 and is divided by 5 upon plateau with a minimum delta of 0.01, and a patience of 8 epochs. The training is automatically stopped when the LR is reduced for the fourth time. For the evaluation on the testing set, the selected epoch is the one among the later half of the epochs, that shows the best pixel-level accuracy on the validation set.

For all active learning methods tested in this work, the initial models are trained again on the union of their initial training dataset and the newly selected tiles. To improve the \emph{weak} model, the labelling budget is set at 1,000 tiles while for the \emph{strong} model the budget is set at 5,000 tiles.

\subsection{Pool of unlabelled images}

The pool of unlabelled images also consists of images from \emph{Maxar} based on their acquisition date. Only images acquired between 2017 January 1st and 2017 July 1st are selected. The images used to train the initial \emph{weak} and \emph{strong} models have mostly civilian aircraft and the images in the unlabelled pool contain more military aircraft. Table~\ref{tab:initial_models} gives the number of aircraft and tiles available in the pool (available as we have access to the labels of the whole pool but do not use them for the selection method).

\begin{table}[bt]
\centering
\caption{Data used to train the \emph{weak} and \emph{strong} initial models and pool of available data.}\label{tab:initial_models}
\begin{tabular}{|c|c|c|c|c|c|}
\hline
Usage & Period & Aircraft & Tiles\\
\hline
Weak model training & 2010-2012  & 951 & 2,740 \\
Strong model training & 2010-2017  & 67,008 & 54,851\\
\hline
Unlabelled pool & 2017   & 11,017 & $\sim$~400k \\
\hline
\end{tabular}
\end{table}

\subsection{Test set and metrics}

Performances are evaluated on a testing set composed of 30 full-size satellite images from 16 different locations that contains mostly military aircraft among 1,532 aircraft in total.
At inference, the segmentation maps obtained for aircraft class are thresholded and the remaining pixel clusters are vectorized to obtain polygons.
Precision-recall curves are used to assess the performances of the models. An aircraft in the ground truth is considered as a true positive if at least half of its area is covered by a predicted segmentation polygon. If a predicted polygon does not cover at least half of the area of an aircraft in the ground truth, it is considered as a false positive.  The precision-recall curves are obtained by varying the threshold applied to the segmentation map. 

\subsection{Results}

As in \cite{caid}, the two proposed hybrid approaches are compared to two baselines. For the baseline, called \emph{unlimited}, the initial model is trained again with a labelling budget considered as unlimited: all the images in the unlabelled pool are labelled and added to the initial training dataset. More precisely, all the tiles containing aircraft are used along with randomly selected negative tiles (10\% of the number of tiles with aircraft). In the end, 5700 tiles are added to the initial dataset. This baseline is the only model for which the pre-selection step is not performed. The second baseline, called \emph{random}, uses the same labelling budget as the active learning approaches for \emph{weak} and \emph{strong} models but the newly added tiles are randomly selected after the pre-selection step.

\begin{figure}[bt]
    \centering
    \includegraphics[width=6.4cm]{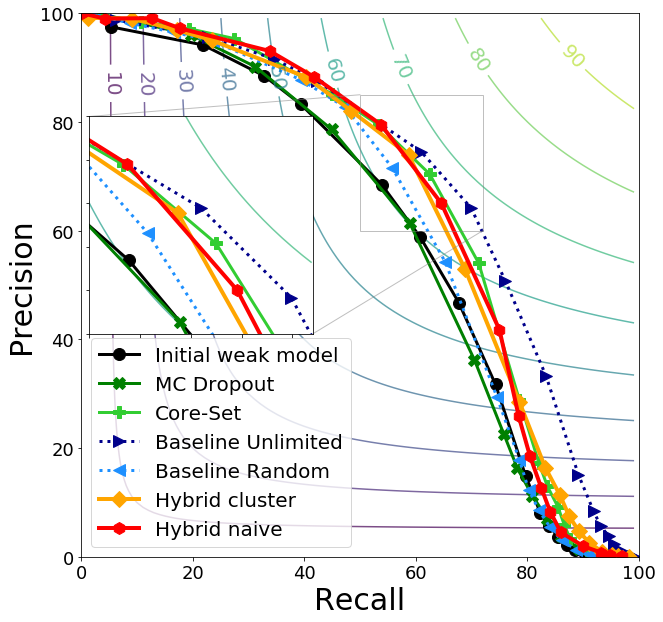}
    \includegraphics[width=6.4cm]{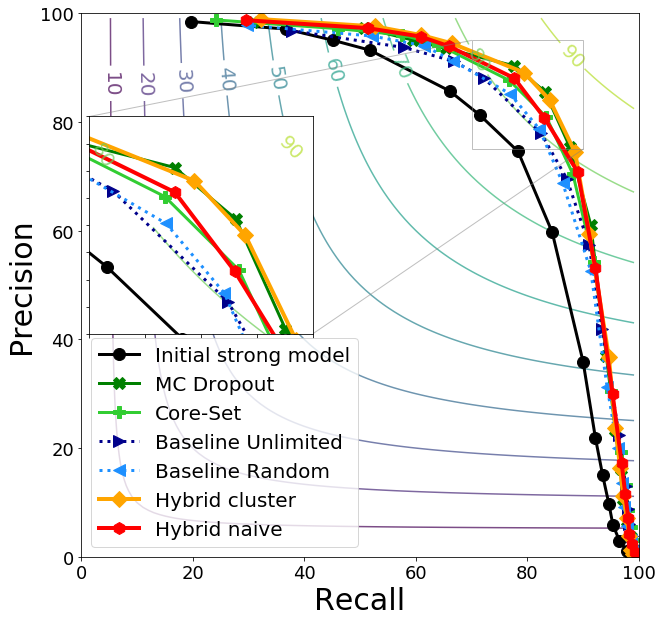}
    \caption{\label{fig:perfo}%
   Performance of the two hybrid active learning methods compared to the results obtained with the methods from \cite{caid} and with the baselines. Top: \emph{weak} model. Bottom: \emph{strong model}.}
\end{figure}

Figure~\ref{fig:perfo} shows the results. 
For the \emph{weak} model, both of the hybrid methods are competitive with the Core-Set method and have performances far above the MC dropout method.
For the \emph{strong} model, the hybrid clustering method provides better results than the naive one. Additionally, hybrid clustering reaches the performances of the MC dropout method while surpassing the \emph{unlimited} baseline. These results show that in both cases, the hybrid clustering is competitive with the best of the two initial active learning methods: MC dropout and Core-Set. Hybrid clustering produces results that are agnostic with respect to the initial strength of the model. It allows to reach high performances without having to make any assumption on the performance of the initial model. This is a valuable asset compared to the methods presented in \cite{caid} when deploying operational models.

\section{Conclusion and perspectives}
\label{conclusion}

We propose a novel hybrid clustering method that leverages both
diversity and uncertainty by combining the MC dropout and Core-Set methods. For both weak and strong initial models, the hybrid clustering method shows a performance gain that is competitive with both the initial uncertainty and diversity methods. Hybrid clustering presents the huge advantage of being agnostic with respect to the performance of the initial model to be updated with new data. This was not the case with the methods we previously proposed in \cite{caid}.

Future developments could include applying these techniques along with continuous learning methods to avoid re-using initial data. This is all the more interesting that operational detectors do not always have access to their initial training data when they are deployed.

\textit{This research was pursued through Accelerator Contract No DSTLX1000148878. We thank DSTL and DASA for this very fruitful collaboration, enlightening discussions and suggestions.}

% \vfill
% \pagebreak

% References using biblio.bib
\bibliographystyle{IEEEbib}
\bibliography{biblio}

\end{document}